\documentclass[sigconf]{acmart} %anonymous

\usepackage{booktabs} % For formal tables
\usepackage[figurename=Fig.]{caption}
\usepackage{multicol}
\usepackage{multirow}
\usepackage{fancyhdr}
\usepackage{balance}
\usepackage{comment}
\usepackage{placeins}
\usepackage{graphicx}
\usepackage{float}
\setcopyright{rightsretained}

\AtBeginDocument{%
  \providecommand\BibTeX{{%
    \normalfont B\kern-0.5em{\scshape i\kern-0.25em b}\kern-0.8em\TeX}}}

\copyrightyear{2021}
\acmYear{2021}
\setcopyright{acmcopyright}\acmConference[AdvM '21]{Proceedings of the 1st International Workshop on Adversarial Learning for Multimedia}{October 20, 2021}{Virtual Event, China}
\acmBooktitle{Proceedings of the 1st International Workshop on Adversarial Learning for Multimedia (AdvM '21), October 20, 2021, Virtual Event, China}
\acmPrice{15.00}
\acmDOI{10.1145/3475724.3483610}
\acmISBN{978-1-4503-8672-2/21/10}
\begin{document}
\fancyhead{}
\fancyfoot{}

\title{Frequency Centric Defense Mechanisms against Adversarial Examples}

\author{Sanket B. Shah}
\authornote{First co-authors with equal contribution to this research.}

\affiliation{%
  \institution{Ahmedabad University}
%   \streetaddress{XYZ}
  \city{Ahmedabad}
  \state{Gujarat}
  \country{India}
%   \postcode{000000}
}
\author{Param Raval}
\authornotemark[1]
\affiliation{%
  \institution{Ahmedabad University}
%   \streetaddress{XYZ}
  \city{Ahmedabad}
  \state{Gujarat}
  \country{India}
%   \postcode{000000}
}
\author{Harin Khakhi}
\authornotemark[1]
\affiliation{%
  \institution{Ahmedabad University}
%   \streetaddress{XYZ}
  \city{Ahmedabad}
  \state{Gujarat}
  \country{India}
%   \postcode{000000}
}
\author{Mehul S. Raval}
\authornote{Corresponding author. Email: mehul.raval@gmail.com}
\affiliation{%
  \institution{Ahmedabad University}
%   \streetaddress{XYZ}
  \city{Ahmedabad}
  \state{Gujarat}
  \country{India}
%   \postcode{000000}
}
% \renewcommand{\shortauthors}{}

\begin{abstract}
Adversarial example (AE) aims at fooling a Convolution Neural Network by introducing small perturbations in the input image. The proposed work uses the magnitude and phase of the Fourier Spectrum and the entropy of the image to defend against AE. We demonstrate the defense in two ways: by training an adversarial detector and denoising the adversarial effect. Experiments were conducted on the low-resolution CIFAR-10 and high-resolution ImageNet datasets. The adversarial detector has 99\% accuracy for FGSM and PGD attacks on the CIFAR-10 dataset. However, the detection accuracy falls to 50\%  for sophisticated DeepFool and Carlini \& Wagner attacks on ImageNet. We overcome the limitation by using autoencoder and show that 70\% of AEs are correctly classified after denoising.
\end{abstract}




\maketitle

\section{Introduction}
Convolutional Neural Networks (CNNs) have shown robust performance in object detection, classification, pose estimation, and image understanding \cite{goodfellow2015explaining}.  However, introducing imperceptible perturbations leads the classifier to yield incorrect results. Such manipulation of an image is called an adversarial attack(AA), and the perturbed image is called an adversarial example.
\begin{comment}
An AE, when used in the real world, can prove to be perilous to automated systems like face recognition and human detection and perhaps fatal in the case of self driving cars \cite{eykholt2018robust}.

There are two kinds of methods to generate adversarial attacks - black-box methods, which assume no prior knowledge about the model and its parameters. Whereas, the white box method intelligently uses the information regarding the model’s architecture to introduce perturbations. This information gives white box methods an edge over the black box methods in successfully fooling the model. 
\end{comment}
White box attacks such as Fast Gradient Sign Method (FGSM) \cite{rozsa2016adversarial} and Projected Gradient Descent (PGD) \cite{madry2017towards} exploit the linear nature of models to introduce noise to the images. More sophisticated attacks such as DeepFool \cite{moosavi2016deepfool}, and Carlini \& Wagner (CW) \cite{carliniwagner} introduce just the right amount of noise to fool the model.
%%\vspace{-2mm}
%%\subsection{Existing Body of Knowledge}
\begin{comment}
There are two kinds of countermeasures against AA: 1) reactive - detect adversarial examples after deep neural networks are built; 2) proactive - make deep neural networks more robust before adversaries generate AEs \cite{yuan2019adversarial}. 
\end{comment}

Adversarial training is a proactive method that trains models with AEs and their benign counterparts.
\begin{comment}
This approach of showing the classifier both the images would allow it to learn how to correctly classify the AE, albeit at the cost of training on many diverse AEs. 
\end{comment}
The transformation of inputs like feature squeezing \cite{DBLP:FeatureSqueezing} also nullifies the effect of the AA. Though it is computationally cheaper, it is not robust against sophisticated white-box attacks on high-resolution images \cite{defensenotstrong}.
\begin{comment}
Methods such as Principal Component Analysis (PCA) \cite{zhang2020principal} and feature squeezing \cite{DBLP:FeatureSqueezing} transform the input image to nullify the effect of the adversarial perturbation. However, these methods are not as robust against sophisticated white-box attacks and spatially rich, real-world images \cite{defensenotstrong, aedreview, sharma2018bypassing}. 
In order to detect an AE, we take a cue from adversarial training and train binary classifier on benign images and AE. 
\end{comment}
Moreover, a simple CNN is unable to capture the subtle perturbations in the case of attacking real-world images with DeepFool \cite{moosavi2016deepfool}, and CW \cite{carliniwagner}. It has been observed that perturbations of AA are more amplified in the frequency domain. Dong et al. \cite{DBLP:fourier_lin} show that the classifier trained with low pass filter (LPF) feature squeezing detects high-frequency noise addition. However, robustness degrades for low-frequency corruptions and perturbations.
Harder et al. \cite{harder2021spectraldefense} use Fourier spectrum projections for adversarial training. They show that CNN is inherently sensitive to Fourier basis functions, and AA would also be represented better in the Fourier domain.
\begin{comment}
The degree of randomness over local neighborhood is also a measure of the spatial frequency and it is characterized by entropy of the image. Hence, a combination of Fourier transform and entropy is exploited in the proposal to extract frequency information about the image.
Experiments conducted on the ImageNet \cite{deng2009imagenet} dataset revealed that the proposed frequency-centric approach has a low AE detection rate against DeepFool and CW. After careful analysis, we propose a solution that involved denoising the adversarial images, which attempts to restore the AE to its benign form.
\end{comment}

%%\subsection{Contributions of the Proposal and Paper Organization}
As a contribution to the proposal, we introduce the idea to use entropy, a measure of spatial frequency and randomness over the local neighborhood. Thus entropy, Magnitude of Fourier Spectrum (MFS), and Phase of Fourier Spectrum (PFS) generate comprehensive features with spatial and frequency representation. 
\begin{comment}
By transformation of images to a 4D tensor and processing by a 3D CNN architecture, the model provides state-of-the-art accuracy against FGSM and PGD attacks. 
\end{comment}
We point out the shortcoming of adversarial training against DeepFool \cite{moosavi2016deepfool}, and CW \cite{carliniwagner}, for ImageNet. We further observe that denoising the AE can help nullify the perturbations \cite{chen2018comparative}. We showcase the use of an autoencoder to filter the adversarial noise.
\begin{comment}
%%\subsection{Paper Organisation}
The remaining paper is organized as follows. Section \ref{sec:adv_attacks} discusses attacks and dataset generation, while Section \ref{sec:feature} explains image features. The proposed frequency-centric AE detection is in Section \ref{sec:ae_detect}, while AE denoising is uncovered in Section \ref{sec:ae_denoise}. Section \ref{sec:similarity} uses similarity metrics to point out the effectiveness of various feature projections in capturing the presence of perturbations. Section \ref{sec:conclusion} concludes this work and discusses the future work.
\end{comment}
\section{Adversarial Attacks}
\label{sec:adv_attacks}

As stated in \cite{goodfellow2015explaining} adversarial examples are inputs formed by applying small but intentional perturbations to samples from the dataset. The perturbed input results in the model predicting an incorrect answer with high confidence. We consider examples from four attack methods: FGSM, PGD, DeepFool, and CW to train and validate the proposed scheme. In addition to being popular in the latest literature, these four attack methods help in spanning a range of complexities in the attack generation, with FGSM being an early yet weaker attack method, PGD being an advancement of the former. In contrast, the attacks generated by DeepFool and CW are more sophisticated.

\subsection{Types of Adversarial Attacks}

\subsubsection{FGSM\nopunct}
 \cite{rozsa2016adversarial} - The method attempts to exploit the linearity of models and create cheap, analytical perturbations of linear models that can damage neural networks. It can be described in one step as follows.
\begin{equation}
  X_{adv} = X + \epsilon\cdot sign(\nabla_X J(\theta, X, y))
\end{equation}

where, $\theta$ = parameters of model, $X$ = input to the model, $y$ = associated label, $J()$ is the loss function, $\nabla$ is gradient, and $\epsilon$ controls the perturbation size. FGSM is designed to be fast and it does not focus on strategic insertion of perturbations.

\subsubsection{PGD\nopunct}\cite{madry2017towards}
 - This method is a multi-step variant of the FGSM technique. Instead of taking one step in the gradient direction, multiple small steps are taken, and hence it is a more sophisticated version of the FGSM.

\subsubsection{DeepFool\nopunct} \cite{moosavi2016deepfool}
 - This method computes the minimal perturbations required to misclassify an example by projecting the sample onto a hyperplane and thus finding the closest decision boundary. It is based on an iterative linearization of the classifier to generate minimal perturbations that are sufficient to change classification labels.

\subsubsection{CW\nopunct} \cite{carliniwagner}
 - It has three optimization variations for attack generation, L$_0$, L$_{\infty}$, and L$_2$, we use the most popular L$_2$ variant. For a given image $X$, CW generates an adversarial example ${X_{adv}}$ by solving the following L$_2$ optimization
problem:
\begin{equation}
  \min \left \| \frac{1}{2}(\tanh({X_{adv}}) + 1) - X \right \|_{2}^{2} + c\cdot f(\frac{1}{2}(\tanh({X_{adv}}) + 1)))
\end{equation}  
  where, c is a constant that determines the power of the attack and \textit{f} is defined as,
\begin{equation}
  f(x') = \max (\max\{Z(x')_{i}: i \neq t\} - \max(Z(x')_{t}), 0)
\end{equation}

\textit{Z(x)} represents the output of all layers except the softmax. It is observed and later demonstrated that this method is more robust and strategic in terms of insertion of perturbations as compared to the other attack methods.

\begin{figure}[]
    \includegraphics[width=1\columnwidth]{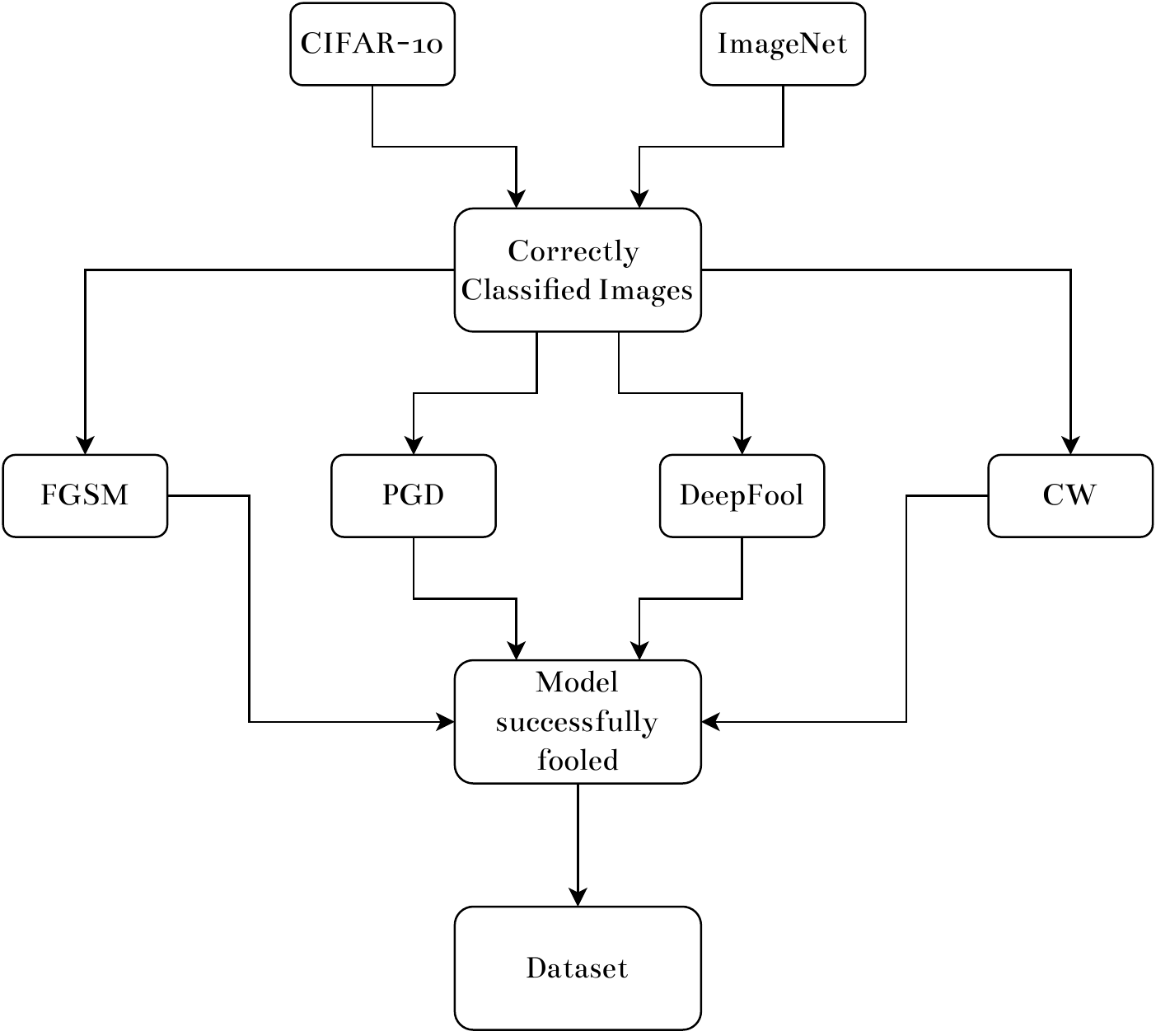}
    \caption{Generation of AE dataset.}
    \label{fig:dataset}
\end{figure}

\subsection{Adversarial Example Dataset Generation}

Two datasets were considered for conducting the experiments described in this work. The first dataset is a portion of the CIFAR-10 \cite{krizhevsky2009learning}  that contains ten classes (airplanes, cars, etc.) and 8000 images with dimensions 32x32x3. The second dataset is derived from the ImageNet \cite{deng2009imagenet}, containing ten classes (parachute, chainsaw, etc.) and 1000 high resolution images of the size 299x299x3. Thus proposed methods are applicable to real world images. For a single image,  five variants of the same image exist in our final dataset - one benign image and the other four being the AEs. The process of obtaining the AE dataset is as described in Fig. \ref{fig:dataset}. Images from CIFAR-10 and ImageNet are classified using pretrained models of VGG16 \cite{vgg16} and InceptionNetV3 \cite{DBLP:InceptionNetV3} respectively. We chose the popular VGG16 architecture for CIFAR-10 to fairly compare our results with \cite{harder2021spectraldefense}. The model is trained on the CIFAR-10 training set, achieving 83.26\% test accuracy \footnote{https://github.com/kuangliu/pytorch-cifar}. However, since the samples in ImageNet are of higher resolution and complexities, InceptionNetV3 was chosen to obtain a better classification rate on benign images. InceptionNetV3 has 42-layer deep architecture and has achieved an error rate of 5.6\% \cite{DBLP:InceptionNetV3} after training on more than a million samples from ImageNet. With a more robust classification model, attack generation, detection as well as denoising becomes more challenging.  

The image for which the class is unambiguously predicted with high confidence was shortlisted to be included in the AE set. Each of the four white-box attacks are performed with respect to the base model: VGG16 for CIFAR-10 and InceptionNetV3 for ImageNet. Next, we retain the AEs that can fool the classifier with good confidence. The perturbation size ($\epsilon$) values for the attacks are set to retain good perceptual fidelity yet fool the network successfully. We use $\epsilon$=0.03 for FGSM and PGD, and set maximum iterations to 1,000 for DeepFool and CW as suggested by \cite{harder2021spectraldefense}. 

\begin{figure*}[!h]
  \centering
  \includegraphics[width=\textwidth]{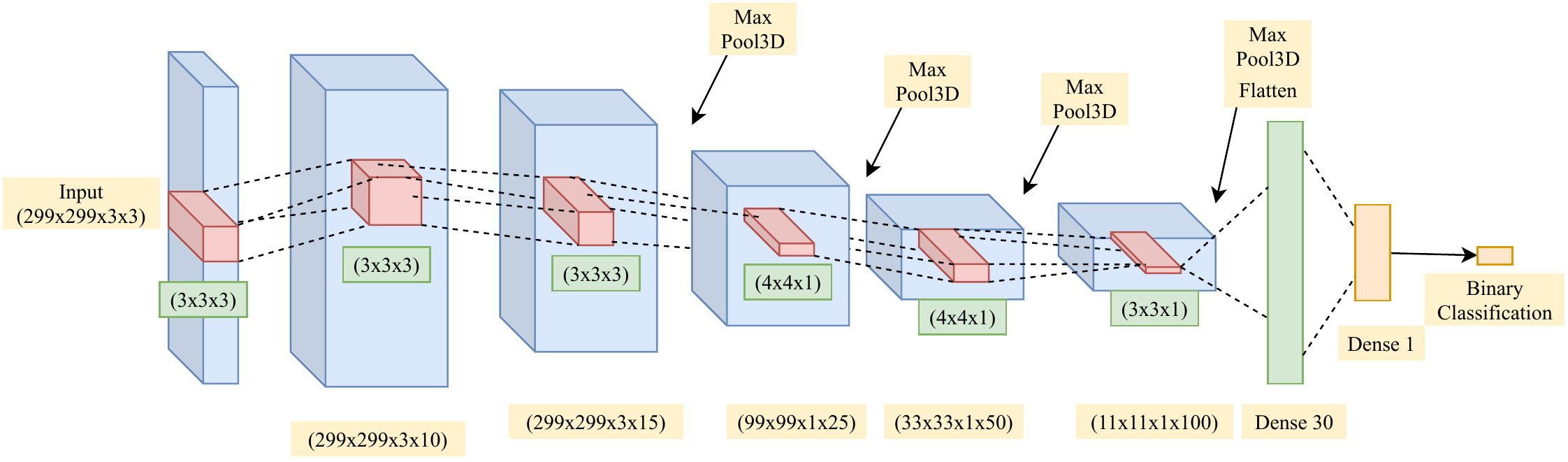}
  \caption{Architecture of 3D CNN adversarial training detector for ImageNet.}
  \label{fig:AEdetect}
\end{figure*}

\begin{figure}[]
    \includegraphics[width=0.8\columnwidth,height=0.8\columnwidth]{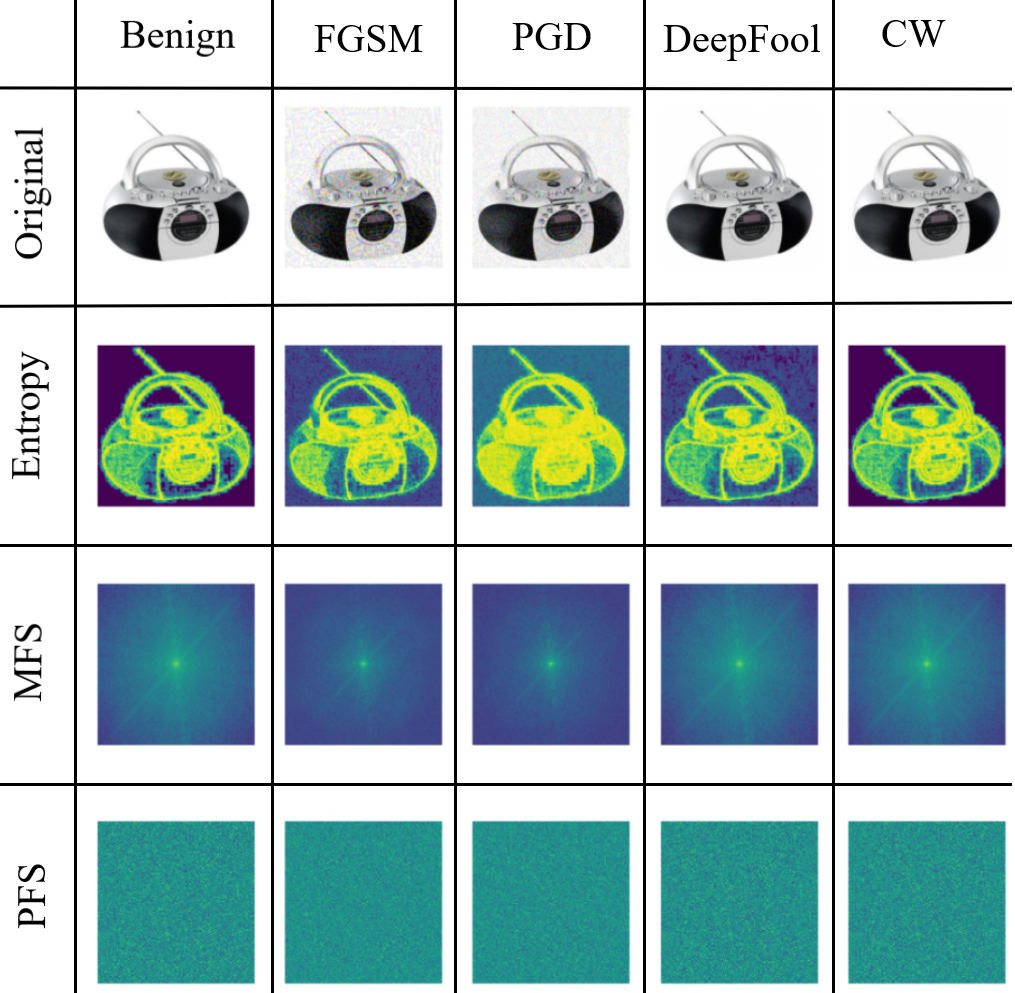}
    \caption{Entropy, MFS and PFS representations for benign and AE images.}
    \label{fig:features}
\end{figure}

\section{Feature Selection}
\label{sec:feature}

The AEs obtained through white-box methods do not display an apparent deviation from their benign counterparts in the spatial domain as the perturbations are visually imperceptible. As seen in Fig. \ref{fig:features}, AEs generated by DeepFool and CW are visually similar to the benign image. In contrast, FGSM and PGD retain all of the visual features of the benign image but with some perceptible noise. However, frequency transforms focusing on specific features of an image can magnify these differences. More specifically, CNNs are more sensitive to Fourier basis functions \cite{DBLP:tsuzuku-sato-ft-sensitivity}. Thus, perturbations made by white-box attacks generated are also likely to be characterized better with Fourier transform.
\begin{comment}
A representation of various frequencies present in an image can be obtained by visualizing it in the Fourier domain. For an image channel $X \in [0,1]^{N \times N}$, the 2D Discrete Fourier Transform (DFT) can be described
as follows:
\begin{equation}
  \mathcal{F}(X)(l,k) = \sum_{n,m=0}^{N} e^{-2\pi\iota\frac{lm+kn}{N}}X(m,n),
\end{equation}
for $l,k$ in [$0$, $N-1$].

DFT was performed on the benign images and AE to obtain the MFS and PFS projections of the corresponding image. MFS can be described as:
\begin{equation}
  \left| \mathcal{F}(X)(l,k) \right| = \sqrt{\textt{Re}(\mathcal{F}(X)(l,k))^{2} + \textt{Im}(\mathcal{F}(X)(l,k))^{2}},
\end{equation}

Similarly PFS can be defined as:
\begin{equation}
  \phi(\mathcal{F}(X)(l,k)) = \arctan\frac{\textt{Re}(\mathcal{F}(X)(l,k))}{\textt{Im}(\mathcal{F}(X)(l,k))}
\end{equation}
\end{comment}

\begin{table*}[]
\caption{\textmd{The hyperparameters for the 3D CNN adversarial training detector.}}
\label{tab:hyperparams}
\centering
 \resizebox{0.95\textwidth}{!}{
\begin{tabular}{|c|c|c|c|c|c|c|c|c|c|c|}
\hline
\multirow{2}{*}{\textbf{Dataset}}    & \multirow{2}{*}{\textbf{Input Shape}} & \multicolumn{6}{c|}{\textbf{Convolutional Layers}}                                                                               & \multicolumn{3}{c|}{\textbf{Dense Layers}}                        \\ \cline{3-11} 
                                     &                                       & \textbf{Hyperparameter} & \textbf{Layer (1)} & \textbf{Layer (2)} & \textbf{Layer (3)} & \textbf{Layer (4)} & \textbf{Layer (5)} & \textbf{Hyperparameter} & \textbf{Layer (1)} & \textbf{Layer (2)} \\ \hline
\multirow{3}{*}{\textbf{CIFAR-10}}   & \multirow{3}{*}{32$\times$32$\times$3$\times$3}                & \textbf{No. of Kernels} & 10                 & 15                 & 25                 & \multirow{3}{*}{-} & \multirow{3}{*}{-} & \textbf{No. of Nodes}   & 20                 & 1                  \\
                                     &                                       & \textbf{Kernel Shape}   & 3$\times$3$\times$3              & 3$\times$3$\times$3              & 3$\times$3$\times$1              &                    &                    & \textbf{Dropout Rate}   & 0.4                & 0.2                \\
                                     &                                       & \textbf{Max Pooling}    & -                  & 3$\times$3$\times$3              & 2$\times$2$\times$1              &                    &                    & \textbf{Activation}     & ReLU               & Sigmoid            \\ \hline
\multirow{3}{*}{\textbf{ImageNet}} & \multirow{3}{*}{299$\times$299$\times$3$\times$3}              & \textbf{No. of Kernels} & 10                 & 15                 & 25                 & 50                 & 100                & \textbf{No. of Nodes}   & 30                 & 1                  \\
                                     &                                       & \textbf{Kernel Shape}   & 3$\times$3$\times$3              & 3$\times$3$\times$3              & 4$\times$4$\times$1              & 4$\times$4$\times$1              & 3$\times$3$\times$1              & \textbf{Dropout Rate}   & 0.4                & 0.2                \\
                                     &                                       & \textbf{Max Pooling}    & -                  & 3$\times$3$\times$3              & 3$\times$3$\times$1              & 3$\times$3$\times$1              & 3$\times$3$\times$1              & \textbf{Activation}     & ReLU               & Sigmoid            \\ \hline
\end{tabular}}
\end{table*}

The MFS and PFS projections of AE and the benign images can be seen in Fig. \ref{fig:features}. It can be observed that the MFS and PFS of the AE has more frequency components as compared to the benign image's magnitude and phase spectrum.  This shows that DFT accentuates the noise. 

The entropy of an image is a measure of the randomness of the pixels in its neighborhood. AEs naturally display more randomness than benign images as the perturbations contribute towards the randomness.  The entropy of the regions affected by perturbations is more than that of the unaffected regions. The entropy projections of benign and AE are shown in Fig. \ref{fig:features} which clearly shows the enhanced entropy for AE. This observation also affirms the understanding that adversarial attacks affect the frequency components of an image.

\section{Detection by Adversarial Training}
\label{sec:ae_detect}
%Explain concept of detection
The AE and benign images form two separate classes, hence making this a binary classification problem. The challenge is to identify the underlying distribution of pixels in benign images and AEs. It is learned with the help of feature projections: MFS, PFS, and entropy for both AEs and benign images.

\subsection{Experimental Setup}
% The dataset uses images with three color bands - RGB. Therefore, the transforms: MFS, PFS, and entropy were performed on all three color channels. In order to use a tensor, the above information is converted to a 4D tensor for every image. Hence, the tensor dimensions are represented by a 4D vector [x,y,f,c] where x, y are the spatial coordinates, f is a feature, and c is a color channel.
The input of this architecture would be a tensor containing feature transforms of benign images and AE: entropy, MFS and PFS. The output of this architecture would be a scalar that represents the probability of an image being AE. The input is of the shape \textit{(x, y, f, c)} where \textit{x},\textit{y} represent the spatial dimensions of the image, \textit{f} represents the feature of the image and \textit{c} represents the RGB color channels. The dataset is split into train and test dataset with a ratio of 70:30 respectively. The CIFAR-10 dataset consists of 8000 images (800 for each of the ten classes) and the ImageNet dataset contains 1000 images (100 for each of the ten classes). 

A self-designed model was built to process the datasets: CIFAR-10 and ImageNet. The model makes use of 3D convolutional filters to exploit the spatial as well as the feature domain of the 4D tensors. The ImageNet dataset contains images of a higher resolution than the images present in the CIFAR-10 dataset. Hence, two variants of the model were created such that the model processing the ImageNet dataset is more profound as shown in Fig. \ref{fig:AEdetect}. Grid search was performed to obtain the best hyperparameters for validation set and they are as shown in Table \ref{tab:hyperparams}. The model was trained with the binary cross-entropy loss and a learning rate of 0.001.

\begin{table}[]
\caption{\textmd{Accuracy / Area Under Curve (AUC) - ROC (\%) for adversarially trained detector method. The attacks are applied on the CIFAR-10 dataset and the VGG-16 architecture.}}
\label{tab:cifardetect}
\resizebox{0.9\columnwidth}{!}{
\begin{tabular}{|c|c|c|c|c|}
\hline
\textbf{Detector}     & \textbf{FGSM} & \textbf{PGD} & \textbf{DeepFool} & \textbf{CW} \\ \hline
Proposed Scheme       & 99.5 / 99.8   & 99.2 / 99.2  & 72.6 / 76.7       & 62.6 / 68.3 \\ \hline
SpectralDefense (InputMFS)\cite{harder2021spectraldefense} & 98.1 / 99.7   & 93.6 / 97.9  & 58.0 / 60.6       & 54.7 / 56.1 \\ \hline
LID\cite{DBLP:lid}                   & 86.4 / 90.8   & 80.4 / 90.0  & 78.9 / 86.8       & 78.1 / 85.3 \\ \hline
MD\cite{mahalanobis}           & 95.6 / 98.8   & 96.0 / 98.6  & 76.1 / 84.6       & 76.9 / 84.6 \\ \hline
\end{tabular}}
\end{table}

\begin{figure}
    \centering
    \includegraphics[width=0.8\columnwidth]{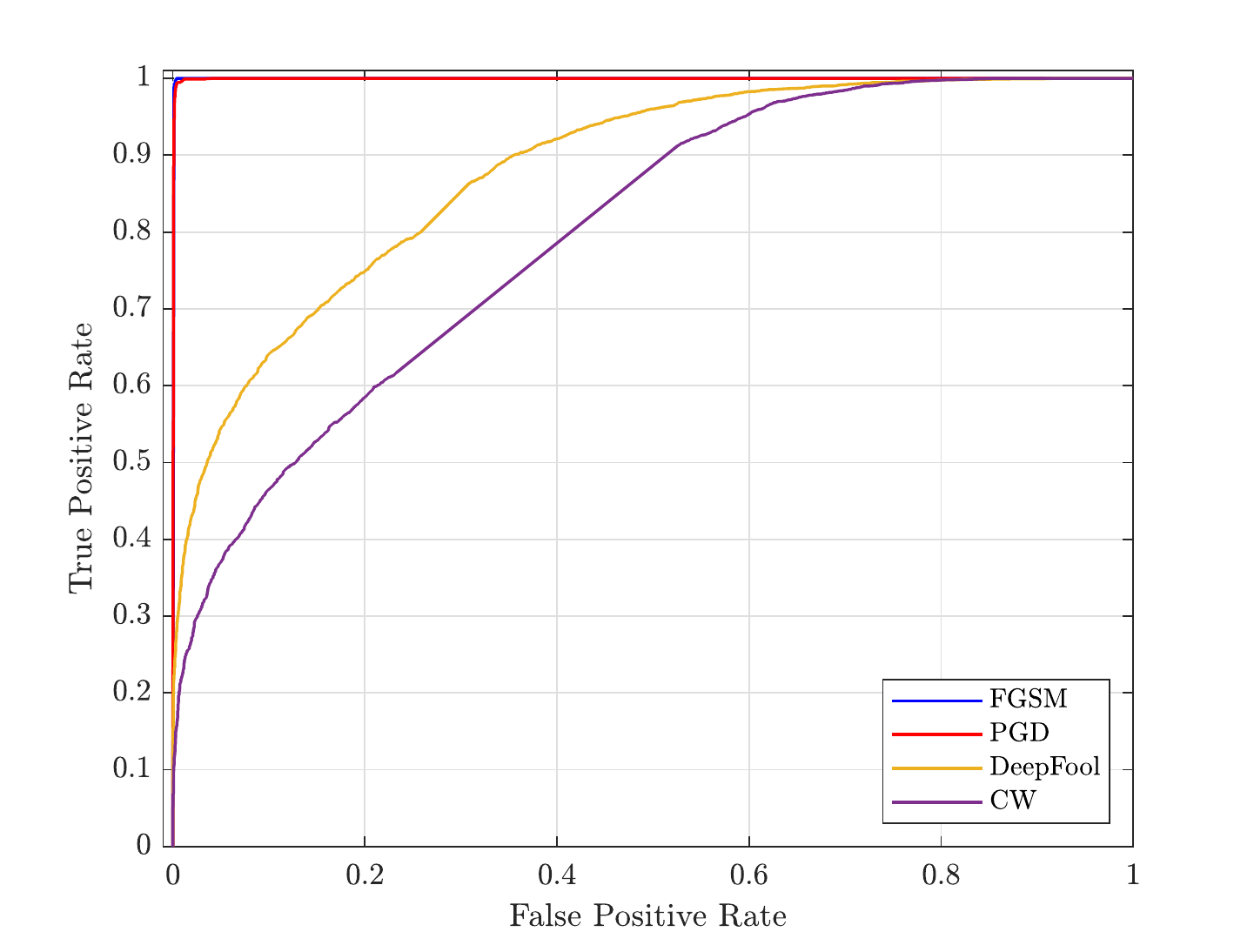}
    \caption{ROC curve for adversarially trained detector on CIFAR-10 dataset.}
    \label{fig:roc}
\end{figure}
\begin{comment}
\begin{table}[]
\centering
\caption{\textmd{The magnitude of perturbations with varying datasets and attack methods.}}
\label{tab:mag_perturb}
\begin{tabular}{|c|c|c|}
\hline
\textbf{}  \textbf{Attack}& \textbf{CIFAR-10} & \textbf{ImageNet} \\ \hline
FGSM     & 1.21$\times$10^{-3}            & 6.66$\times$10^{-4}              \\ \hline
PGD      & 8.21$\times$10^{-4}            & 1.46$\times$10^{-4}              \\ \hline
DeepFool & 9.27$\times$10^{-5}            & 2.55$\times$10^{-5}              \\ \hline
CW       & 6.85$\times$10^{-5}            & 4.50$\times$10^{-6}              \\ \hline
\end{tabular}
\end{table}
\end{comment}
\begin{table}[]
\caption{\textmd{Accuracy (\%) of the proposed adversarially trained detector method on ImageNet.}}
\label{tab:imagenetdetect}
\resizebox{0.9\columnwidth}{!}{
\begin{tabular}{|c|c|c|c|c|}
\hline
\textbf{Attack}   & \textbf{Entropy (Ent.)} & \textbf{MFS}  & \textbf{PFS}  & \begin{tabular}[c]{@{}c@{}}\textbf{Combined}\\ \textbf{(Ent. + MFS + PFS)}\end{tabular} \\ \hline
FGSM     & 95.6    & 91.2 & 93.4 & 98.8                                                                  \\ \hline
PGD      & 93.2    & 91.1 & 90.9 & 96.7                                                                  \\ \hline
DeepFool & 50.7    & 49.9 & 50.3 & 51.8                                                                  \\ \hline
CW       & 49.9    & 50.0 & 50.0 & 50.7                                                                  \\ \hline
\end{tabular}}
\end{table}

\begin{figure*}[h!]
  \centering
  \includegraphics[width=0.9\textwidth]{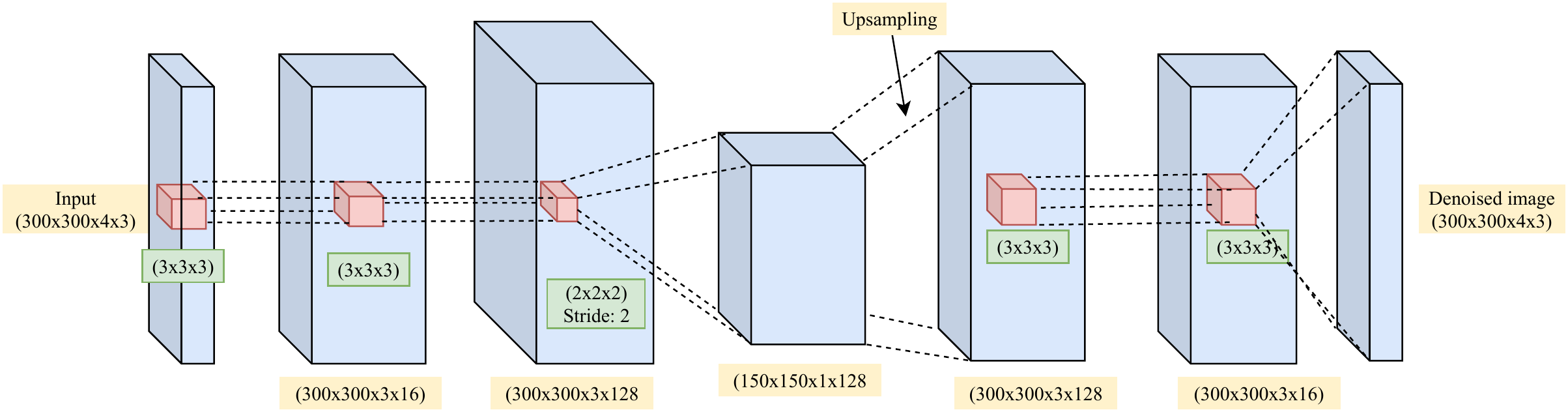}
  \caption{\textmd{Model architecture for denoising the AE.}}
  \label{fig:AEdenoise}
\end{figure*}

\begin{table*}[h!]
\caption{\textmd{The hyperparameters for autoencoder used in denoising.}}
\label{tab:denoise_hyperparams}
\centering
 \resizebox{0.8\textwidth}{!}{
\begin{tabular}{|c|c|c|c|c|c|c|c|c|}
\hline
\textbf{Input Shape}               & \textbf{Hyperparameters} & \textbf{Layer (1)} & \textbf{Layer (2)} & \textbf{Layer (3)} & \textbf{Upsampling}        & \textbf{Layer (4)} & \textbf{Layer (5)} & \textbf{Output Shape}              \\ \hline
\multirow{3}{*}{300$\times$300$\times$4$\times$3} & No. of Kernels           & 16                 & 128                & 128                & \multirow{3}{*}{2$\times$2$\times$1} & 16                 & 3                  & \multirow{3}{*}{300$\times$300$\times$4$\times$3} \\
                                   & Kernel Shape             & 3$\times$3$\times$3          & 3$\times$3$\times$3          & 2$\times$2$\times$2          &                            & 3$\times$3$\times$3          & 3$\times$3$\times$3          &                                    \\
                                   & Stride                   & 1$\times$1$\times$1          & 1$\times$1$\times$1          & 2$\times$2$\times$1          &                            & 1$\times$1$\times$1          & 1$\times$1$\times$1          &                                    \\ \hline
\end{tabular}}
\end{table*}

\subsection{Results and Comparison with Existing Approaches}
We compare the obtained results with three different approaches: 1) SpectralDefense\cite{harder2021spectraldefense}; 2) Local Intrinsic Dimensionality (LID)\cite{DBLP:lid}; 3) Mahalanobis Distance (MD)\cite{mahalanobis}. The comparison of the results on the CIFAR-10 dataset is as shown in Table \ref{tab:cifardetect}. The Receiver Operating Characteristic (ROC) curve for results obtained on detection of attacks of various methods on CIFAR-10 dataset is shown in Fig. \ref{fig:roc}. The time taken in classifying single image with one feature is $6.46 \times 10^{-5}$ seconds for CIFAR-10 and $4.9 \times 10^{-3}$ seconds for ImageNet. The time taken in classifying a single image with all features is $1.35 \times 10^{-4}$ seconds for CIFAR-10 and $1.02 \times 10^{-2}$ seconds for ImageNet.

We observe that entropy maps, which contain local frequency representations, are more sensitive to perturbations made by the attacks and thus add enough information for the detector to learn and outperform the existing approaches. Moreover, using a three-dimensional architecture also contributes to improving results on MFS and PFS as the network has the scope of looking beyond a single channel and it learns inter-channel correlations better for a given transform. We also tested this approach on the ImageNet dataset to ensure applying the proposed technique on real-world, high resolution images. The obtained results are shown in Table \ref{tab:imagenetdetect}. It can be seen that the scheme performs well for FGSM and PGD. However, it fails to detect AE for DeepFool and CW. 

\subsection{Analysis and Discussion}
It is vital to investigate the inability of the proposed approach to confidently detect the ImageNet AE generated by DeepFool and CW. We found that methods like DeepFool and CW introduce the bare minimum perturbations required for fooling a specific model. When the image size increases (in this case, from 32×32 of CIFAR-10 to 299×299 of ImageNet), the magnitude of perturbations per pixel decreases significantly. This leads to increased difficulty in the differentiation of AE from benign images.

In order to track this change in the magnitude of perturbations, we compute the following ratio for all CIFAR-10 and ImageNet images used in the proposed method and report the average value.

\begin{equation}
  \delta = \frac{\sum\left| \hat{I} - I\right|}{\sum I},
\end{equation}

where, $I$ is the array of benign image and $\hat{I}$ is the AE. 
The value of perturbations $\delta$ for both the datasets and various attacks is shown in Table \ref{tab:mag_perturb}. It can be observed that the magnitude of perturbation observed in ImageNet dataset is lesser than in CIFAR-10 dataset for every attack method.

\begin{table}[h!]
\centering
\caption{\textmd{The magnitude of perturbations with varying datasets and attack methods.}}
\label{tab:mag_perturb}
\begin{tabular}{|c|c|c|}
\hline
\textbf{}  \textbf{Attack}& \textbf{CIFAR-10} & \textbf{ImageNet} \\ \hline
FGSM     & 1.21$\times$10$^{-3}$            & 6.66$\times$10$^{-4}$              \\ \hline
PGD      & 8.21$\times$10$^{-4}$            & 1.46$\times$10$^{-4}$              \\ \hline
DeepFool & 9.27$\times$10$^{-5}$            & 2.55$\times$10$^{-5}$              \\ \hline
CW       & 6.85$\times$10$^{-5}$            & 4.50$\times$10$^{-6}$              \\ \hline
\end{tabular}
\end{table}

\subsubsection{Similarity Metrics}
\label{subsec:similarity}
We employ cosine similarity, peak signal-to-noise ratio (PSNR),  structural similarity index measure (SSIM), and ERGAS \cite{Wald} metrics to further emphasize the sensitivity of Fourier transform projections to adversarial attacks. These metrics are used to compare: 1. the benign image and its AE; 2.  MFS of benign and AE; 3. the PFS between the two. The ideal score for cosine similarity and SSIM is 1.0, for PSNR is infinity, and ERGAS is 0.0. Table~\ref{tab:cosinesim}, ~\ref{tab:psnr}, and ~\ref{tab:ergas} show the average similarity score for 10 sample images from the ImageNet dataset with their adversarial counterparts.

% \begin{figure*}[]
%   \centering
%   \includegraphics[width=0.9\textwidth]{autoencoder_1.pdf}
%   \caption{\textmd{Model architecture for denoising the AE.}}
%   \label{fig:AEdenoise}
% \end{figure*}
It can be observed in Table~\ref{tab:cosinesim} that cosine similarity is unable to capture the difference as strongly as the other two scores. Similarly SSIM also proved ineffective in sensing the differences. From Table~\ref{tab:psnr} and Table~\ref{tab:ergas} we observe that the scores of MFS and PFS are more sensitive and better in capturing the difference between benign and AE for four attacks. It highlights the sensitivity of Fourier transform to adversarial attacks and shows that these projections contain information vital in differentiating an AE from a benign image. By looking at the ideal values, we see that the similarity scores are superior for DeepFool or CW than FGSM or PGD showing that former attacks produce lower perturbation in the AE.\\

\begin{table}
\caption{\textmd{Cosine similarity for four white-box attacks.}}
  \setlength{\tabcolsep}{0.2\tabcolsep}
  \label{tab:cosinesim}
  \centering
  \resizebox{0.7\columnwidth}{!}{%
  
\begin{tabular}{| c | c | c | c | c |}
\hline
\textbf{Entities/Attack} & \textbf{FGSM} & \textbf{PGD} & \textbf{DeepFool} & \textbf{CW} \\ \hline
Original            & 0.998         & 0.999        & 0.999             & 1           \\ \hline
MFS                      & 0.856         & 0.962        & 0.999             & 0.999       \\ \hline
PFS                      & 0.723         & 0.899        & 0.998             & 0.999       \\ \hline
\end{tabular}%
}
\end{table}

\begin{table}
\caption{\textmd{PSNR for four white-box attacks.}}
  \setlength{\tabcolsep}{0.2\tabcolsep}
  \label{tab:psnr}
  \centering
  \resizebox{0.7\columnwidth}{!}{%
\begin{tabular}{| c | c | c | c | c |}
\hline
\textbf{Entities/Attack} & \textbf{FGSM} & \textbf{PGD} & \textbf{DeepFool} & \textbf{CW} \\ \hline
Original   & 30.63         & 40.07        & 51.14             & 51.12       \\ \hline
MFS             & 12.80         & 14.39        & 30.44             & 41.42       \\ \hline
PFS             & 44.42         & 49.31        & 66.86             & 75.53       \\ \hline
\end{tabular}%
}
\end{table}

\begin{table}[t]
\caption{\textmd{ERGAS for four white-box attacks.}}
  \setlength{\tabcolsep}{0.2\tabcolsep}
  \label{tab:ergas}
  \centering
  \resizebox{0.7\columnwidth}{!}{%
\begin{tabular}{| c | c | c | c | c |}
\hline
\textbf{Entities/Attack} & \textbf{FGSM} & \textbf{PGD} & \textbf{DeepFool} & \textbf{CW} \\ \hline
Original   & 3861.60                        & 1192.84                       & 20.65                              & 3.72                         \\ \hline
MFS             & 254455.52                      & 94393.13                      & 2305.83                            & 618.19                       \\ \hline
PFS            & 36840.43                       & 21453.79                      & 1330.12                            & 276.22                       \\ \hline
\end{tabular}%
}
\end{table}
\begin{comment}
However, training on feature maps to avoid feature adversaries\cite{adaptiveattacks} and exploiting complexity of the 3D convolutional architecture might open better scopes to make the proposed method defend against these attacks.
\end{comment}

\section{Denoising of Adversarial Examples}
\label{sec:ae_denoise}

The benign and AE images can be seen as two points in N-dimensional vector space. Lower perturbation by DeepFool and CW means that AE resides in the close vicinity of the benign image in the vector space. The detector, which relies on a significant distance between AE and benign image for its prediction, will fail when the distance between them is meager or overlaps. It is a drawback for the proposed adversarial training detector while tackling the DeepFool and CW attacks.

The use of autoencoders against adversarial examples is a common and useful approach \cite{chen2018comparative}. An autoencoder can be used in denoising an AE \cite{creswell2018denoising} where the objective is to restore the class of a benign image. One such use of convolutional auto-encoders to purify AEs has been described in \cite{DBLP:autoencoder}. As shown by \cite{DBLP:autoencoderdr} dimensionality reduction using an autoencoder is also an effective defense against adversarial attacks. The method uses autoencoder to perform denoising of FGSM attack on MNIST dataset.

An autoencoder tries to learn the underlying intricacies of the training data distribution in an attempt to map the input image to the output image (or reconstruct the input to match the output). This method can be useful in mapping benign images and their AEs, along with the frequency transforms, and training the network to learn the underlying vector field. Hence, upon learning, the model would indicate the magnitude of transformation required for each vector field to be mapped to its clean representation. Thus, the output of the denoiser will be a denoised version of the input sample.

It is observed in \cite{chen2018comparative} that the performance of autoencoders in the task of denoising an image increases the most in cases where the magnitude of perturbation ($\epsilon$ of noise) is least. Also, it was observed in Table \ref{tab:mag_perturb} that sophisticated attacks like DeepFool and CW insert minimal magnitude of perturbations to the image. Owing to these reasons, the denoising approach can be considered to be a suitable technique in nullifying the perturbations from these attacks. 

\subsection{Experimental Setup}
The objective is to apply denoising to AE generated by FGSM, PGD, DeepFool and CW using MFS, PFS, and entropy. We conduct this experiment on attack images generated from the ImageNet dataset. 

The input of this architecture would be a tensor containing the AE along with its feature transforms: entropy, MFS and PFS. The output of this architecture would be a tensor of the same shape that would contain a generated image along with its feature transforms.

Grid search was performed to obtain the hyperparameters that performs the best on validation set and they are as shown in Table \ref{tab:denoise_hyperparams}. The design of the model used is shown in Fig. \ref{fig:AEdenoise}. The output of the autoencoder comprises of denoised images and its feature transforms. We need to check effectiveness of denoising by measuring the accuracy of correct classification. Therefore, the denoised images are classified using InceptionNetV3.

\subsection{Results}
Experiments were performed on the images generated by the denoiser to validate its performance. An image would lose its original class when perturbations from an attack are introduced to it. The objective is to see whether the denoiser was successful in restoring the original class of the image or not. 

Table \ref{tab:denoise_results} shows the \% of images whose classes were restored on the ImageNet dataset. We can also observe and compare the utility of each feature and their combinations in Table \ref{tab:denoise_results}. Also, as observed in \cite{harder2021spectraldefense} and \cite{autoattackdetect}, MFS provides enough information to the network to remove the perturbations better than PFS. We observe that on an average 70\% of the images are denoised by the autoencoder across all the four attacks. The time taken in denoising a single image with one feature is $2.27 \times 10^{-2}$ seconds, whereas for all the features it is $3.11 \times 10^{-2}$ seconds.

\begin{table}
  \caption{\textmd{Accuracy (\%) of correct classification by InceptionNetV3 after denoising AE from ImageNet.}}
  \setlength{\tabcolsep}{0.2\tabcolsep}
  \label{tab:denoise_results}
  \centering
%   \resizebox{0.7\columnwidth}{!}{%
\begin{tabular}{|c|c|c|c|c|}
\hline
\textbf{Training Set/Attack} & \textbf{FGSM} & \textbf{PGD} & \textbf{DeepFool}         & \textbf{CW}               \\ \hline
Entropy                                                                     & 78            & 75.6         & 74.8                      & 72.2                      \\ \hline
MFS                                                                         & 78.8          & 77.8           & 77                      & 68.6                      \\ \hline
PFS                                                                         & 69.4             & 67.6            & 73.5                        & 65.5                      \\ \hline
Entropy + MFS + PFS                                   & 80.3             & 79.5            &72.1 &69.2 \\ \hline
\end{tabular}%
% }
\end{table}

\section{Conclusion and future work}
\label{sec:conclusion}
Adversarial examples aim at fooling the model by strategically introducing subtle perturbations in an image. In this work, we study four white-box attacks and observe the effectiveness of Magnitude of Fourier Spectrum, Phase of Fourier Spectrum, and entropy in differentiating benign and AEs. The adversarial training provided state of the art accuracy in detecting AE generated by FGSM and PGD but it fails against DeepFool and CW on high resolution images. The above limitation is removed by using autoencoder for denoising the AEs. The autoencoder is denoising 70\% of the AE across all four attacks. In future we plan to explore other frequency transforms, critically use information present in the CNN layers, and improvise on the architecture. As a norm, defense methods have to be evaluated against adaptive white-box attacks \cite{adaptiveattacks}. We need further experimentation to evaluate robustness of the proposed method against these attacks. 
% In this work, we studied the effectiveness of frequency domain projections in capturing the perturbations caused by adversarial attacks that are imperceptible to the human eye. We build on existing work using entropy maps in tandem with the magnitude and phase spectra of Fourier Transform of benign image and its AE. Using these transformed images we propose a adversarial detector that outperforms existing detectors using similar methods on CIFAR-10. The transformed images are generated from the input images (benign and AE) and trained using a multi-channel three-dimensional CNN. For real-world images, we propose an adversarial denoising strategy using three-dimensional autoencoders trained on the aforementioned transformed projections. We evaluate this method on a subset of the ImageNet dataset using InceptionNetV3 as the base classifier.

% Future work on this promising work would involve using frequency transforms of the intermediate feature maps from the classifier to train the detector and denoiser. Finally, a more complex architecture for the autoencoder can be realised using better computational resources.

% \begin{multicols*}
\bibliographystyle{ACM-Reference-Format}
% \bibliographystyle{unsrtnat}
% \setcitestyle{numbers,sort&compress}
\balance
\bibliography{a_references.bib}
% \end{multicols*}

\end{document}